\documentclass[10pt,twocolumn,letterpaper]{article}

\usepackage[pagenumbers]{cvpr} 

\usepackage{amsmath}
\usepackage{amssymb}
\usepackage{booktabs}
\usepackage{color}
\usepackage{multirow}
\usepackage{graphicx} 
\usepackage[accsupp]{axessibility}

\definecolor{cvprblue}{rgb}{0.21,0.49,0.74}
\usepackage[pagebackref,breaklinks,colorlinks,allcolors=cvprblue]{hyperref}

\def\paperID{2045} 
\def\confName{CVPR}
\def\confYear{2026}

\title{X-WIN: Building Chest Radiograph World Model via Predictive Sensing}

\author{
Zefan Yang$^1$\ \ \ \ Ge Wang$^1$\ \ \ \ James Hendler$^1$\ \ \ \ Mannudeep K. Kalra$^2$\ \ \ \ Pingkun Yan$^1$\thanks{Corresponding author}\\
{$^1$Rensselaer Polytechnic Institute\ \ \ \ \ \ $^2$Massachusetts General Hospital}\\
{\small\texttt{yangz16@rpi.edu, yanp2@rpi.edu}}
}

\begin{document}
\maketitle
\thispagestyle{empty}
\pagestyle{empty}
\begin{abstract}
Chest X-ray radiography (CXR) is an essential medical imaging technique for disease diagnosis. However, as 2D projectional images, CXRs are limited by structural superposition and hence fail to capture 3D anatomies. This limitation makes representation learning and disease diagnosis challenging.
To address this challenge, we propose a novel CXR world model named X-WIN, which distills volumetric knowledge from chest computed tomography (CT) by learning to predict its 2D projections in latent space. The core idea is that a world model with internalized knowledge of 3D anatomical structure can predict CXRs under various transformations in 3D space. During projection prediction, we introduce an affinity-guided contrastive alignment loss that leverages mutual similarities to capture rich, correlated information across projections from the same volume. To improve model adaptability, we incorporate real CXRs into training through masked image modeling and employ a domain classifier to encourage statistically similar representations for real and simulated CXRs. Comprehensive experiments show that X-WIN outperforms existing foundation models on diverse downstream tasks using linear probing and few-shot fine-tuning. X-WIN also demonstrates the ability to render 2D projections for reconstructing a 3D CT volume.


\end{abstract}

\section{Introduction}
\label{sec:intro}
Chest X-ray radiography (CXR) is the most widely-used imaging technique for thoracic disease detection \cite{raoof2012interpretation}. However, as 2D projectional images, CXRs are limited by structural superposition, and hence fail to capture 3D anatomies of various organs. In comparison, chest computed tomography (CT) is a medical imaging technique which similarly measures the X-ray attenuation by different tissues, but can obtain detailed 3D internal structure via tomographic reconstruction \cite{suetens2017fundamentals}.
However, CT scanning is nearly five times the cost of CXR \cite{flores2017ma03}. CT also delivers notably more radiation to patients compared to CXR, which can potentially cause adverse health effects. Moreover, CT accessibility is considerably more restricted compared to CXR, especially in underdeveloped regions \cite{awolowo2024critical}, due to high installation and maintenance costs. The tradeoffs between rich spatial information on CT and safety, cost-effectiveness, and accessibility of CXR motivate us to devise a deep learning method that can learn spatial knowledge from CT to attain enhanced disease diagnosis on CXRs. 

\begin{figure}
    \centering
    \includegraphics[width=\linewidth]{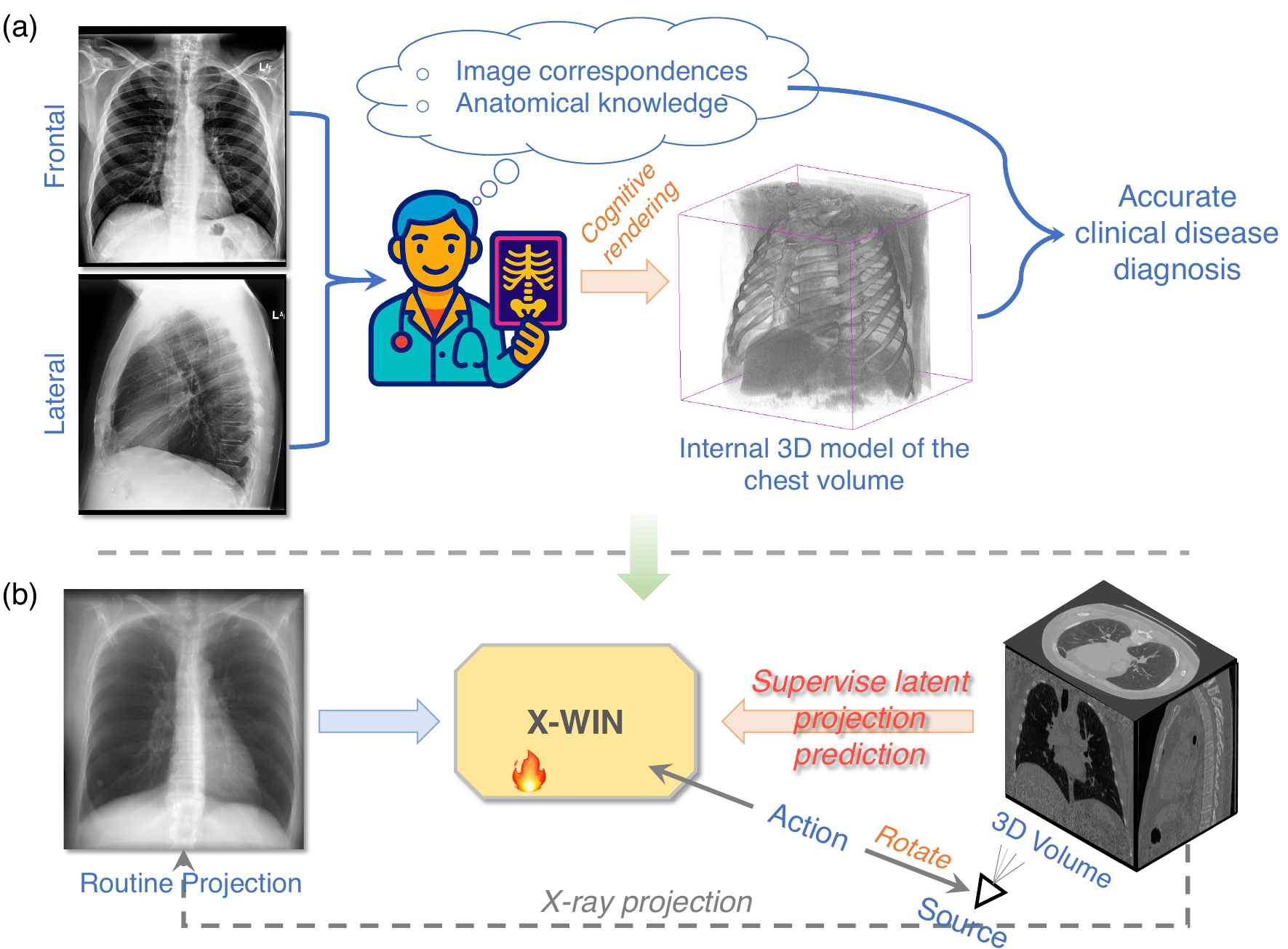}
    \caption{\textbf{Illustration of our motivation.} (a) Given routine frontal and lateral CXRs, radiologists can reconstruct a 3D model of the chest volume based on their anatomical knowledge. This capability helps them make diagnostic decisions despite the occlusion of structures. (b) Inspired by our observation in (a), we propose a novel X-ray World Intelligence Network (X-WIN) that builds a world model by predicting new projection views under various transformations of an X-ray source.}
    \label{fig:teaser}
    \vspace{-10pt}
\end{figure}

Recently, world modeling \cite{hafner2025dreamerv3, bardes2024revisiting, assran2025v, bar2025navigation, chen2025planning} has emerged as a technique to embed an internal understanding of the world into an agent model, which learns meaningful visual representations that capture spatial dynamics of the environment. Such representations can improve scene understanding, navigation, and action planning. In the field of medical imaging, Yang \textit{et~al.} recently introduced CheXWorld \cite{yue2025chexworld}, a world model for CXR interpretation, which devises action-conditioned self-supervised objectives to learn visual representations of radiographs. However, CheXWorld remains limited to 2D and does not embed any 3D knowledge into the world model. 
Such an absence of 3D information leads to a substantial capability gap between the current CXR world model and radiologists. Based on our observation illustrated in Fig.~\ref{fig:teaser}, radiologists can cognitively reconstruct an internal 3D model of the chest region when perceiving frontal and lateral CXRs. 
To bridge this capability gap, we believe that the CXR world model should capture 3D spatial information and propose a novel method to distill volumetric knowledge from CT.

Specifically, we introduce an innovative X-ray World Intelligence Network (X-WIN), which encodes spatial information from CT by learning to predict its projections in latent space. These projections originate from the same CT volume and characterize 3D structure from different views.
More specifically, our proposed framework consists of two synergistic networks. The first network contains an encoder and a light-weight action-conditioned predictor to generate latent projection prediction. It receives the routine frontal or lateral X-ray projection as input. The second network is an exponential moving average encoder that receives multiple projections of a target CT volume to supervise latent projection prediction. The generation of new projections is controlled by an action to move the X-ray source. This framework matches our goal of distilling volumetric knowledge into a CXR model, as the exponential moving average encoder encodes knowledge from a CT volume and guides the learning of the encoder that processes routine CXRs. We deploy X-WIN on downstream CXR interpretation tasks when the training finishes.

To train the proposed model, we introduce three major losses, including an affinity-guided contrastive alignment loss, a masked image modeling loss (MIM) \cite{assran2023self}, and a structure-preserving domain adaptation loss.
The affinity-guided contrastive alignment loss ensures that the proposed model encodes sufficient nuanced information during projection prediction. It is a softened contrastive alignment loss \cite{oord2018representation} based on mutual similarities to capture rich correlated information across multiple projections.
The MIM loss is incorporated to enable the model to encode local and contextual features and enhance its adaptability to downstream CXR tasks. The MIM loss is applied to both real and simulated CXRs. Such a loss enforces the model to reconstruct local patches on CXRs, enabling it to learn fine-grained features of anatomical structures and disease abnormalities. The structure-preserving domain adaptation loss is incorporated to learn a cohesive representation space between the real and simulated domains. In this loss, simulated-domain representations are encouraged to resemble the real-domain representations under the guidance of a domain classifier, while preserving structural information via patch-level supervision.


In summary, our proposed method, X-WIN, effectively distills knowledge from 3D CT via world modeling for an enhanced application on CXRs. Extensive experiments on cutting-edge CXR models and standard benchmarks demonstrate that the learned representations achieve substantial improvements over existing methods with comparable model capacities. Additionally, X-WIN exhibits an effective capability to render 2D projections for tomographic reconstruction, demonstrating a robust understanding of 3D structures. Our contributions are as follows:
\begin{itemize}
\item We propose the X-WIN framework for radiograph world modeling, making the first approach to integrate 3D spatial knowledge into a CXR world model.
\item We introduce an affinity-guided contrastive alignment objective, which enhances the encoding of discriminative features and exploits rich correspondences in different projections. 
\item The learned representations of our model have been applied to standard CXR interpretation tasks, achieving state-of-the-art performance with linear probing. Our method exhibits effective 3D reconstruction capability.
\end{itemize} 

\begin{figure*}
    \centering
    \includegraphics[width=\linewidth]{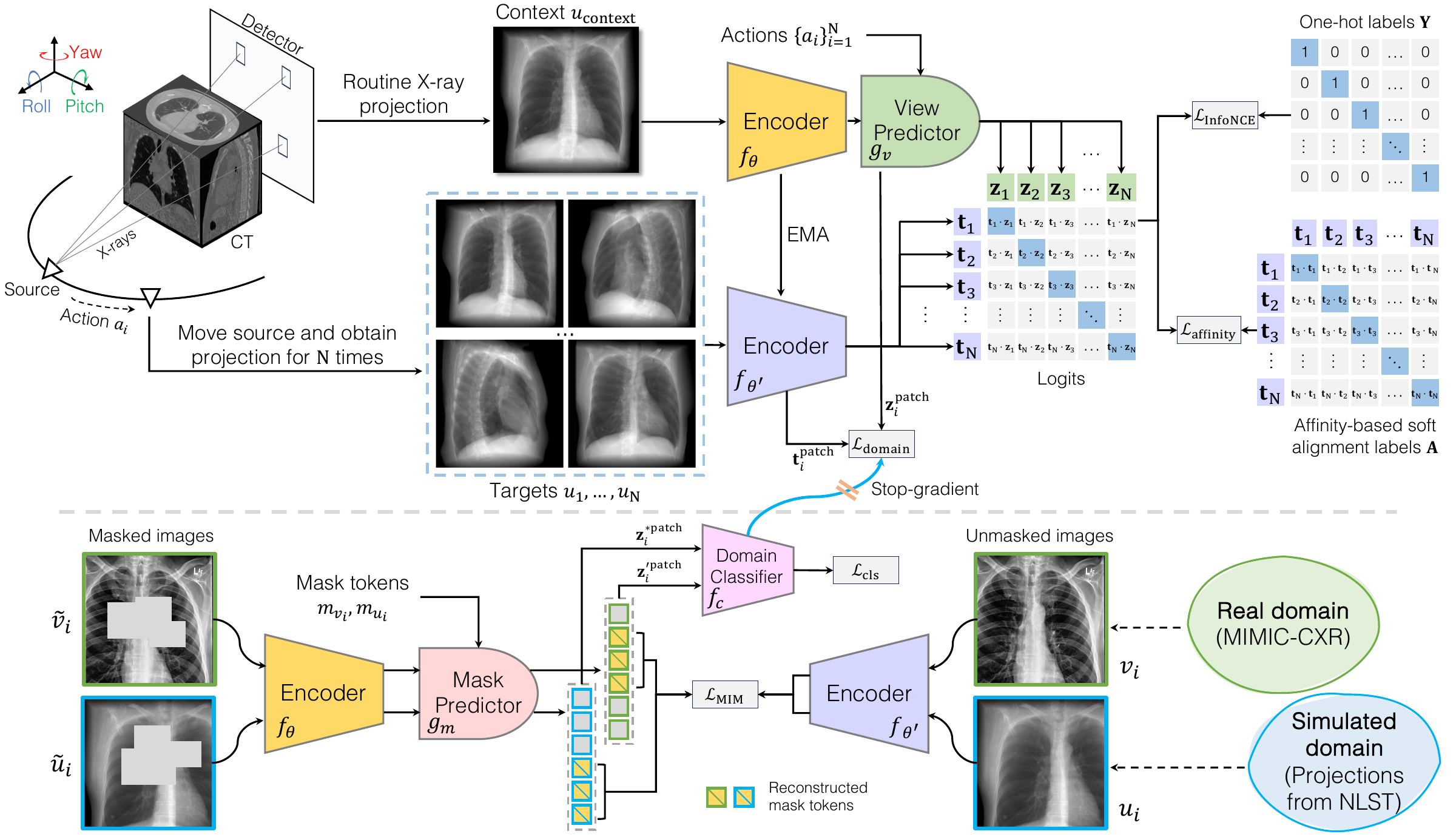}
    \caption{\textbf{Overview of the X-WIN framework.} Top: Our proposed approach distills spatial knowledge from CT by learning to predict its projections in latent space. We generate new projections by rotating an X-ray source controlled by actions. Bottom: Masked image modeling and a domain classifier are incorporated to learn a cohesive representation space between real and simulated domains. EMA denotes exponential moving average.}
    \label{fig:framework}
    \vspace{-10pt}
\end{figure*}

\section{Related Work}
\label{sec:related}
\subsection{World Models}
World models are conceptualized as models that possess implicit spatial cognition of the 3D world \cite{lecun2022path}. In the AI field, the recent Cambrian-S \cite{yang2025cambrian} constructs implicit 3D spatial cognition through \textit{predictive sensing}, where models learn to anticipate their sensory input. This model treats video data as projections of a 3D world and learns to predict the next frame in latent space. It builds implicit 3D spatial cognition into agents, defined as a necessary step towards advanced reasoning. Besides, generative world models \cite{ha2018worldmodels, michelitransformers, bar2025navigation, zuo2025gaussianworld, hafner2025training} learn dynamics of the environment and recursively predict future states in image space. These models estimate rewards given random actions in model-based reinforcement learning, successfully assisting autonomous agent training \cite{hafner2025dreamerv3}. Another line of world models \cite{assran2023self, poudel2024recore, bardesrevisiting, zhoudino, assran2025v, baldassarre2025back, chen2025planning} focus on leveraging visual representations of the environment, motivated by the observation that humans cannot generate photorealistic images yet exhibit superior planning capabilities. In the medical field, recent works propose CheXWorld \cite{yue2025chexworld} that learns representations in 2D radiographs and MeWM \cite{yang2025medical} that models progression of tumors under different treatments. Different from CheXWorld that learns 2D local features and global geometry, our method integrates 3D spatial knowledge by predicting new projections under various 3D transformations.


\subsection{Knowledge Distillation}
Knowledge distillation \cite{hinton2014distilling} has been proposed to compress knowledge in a teacher model into a student model \cite{tarvainen2017mean, he2020momentum, grill2020bootstrap, oquab2024dinov2}. In world modeling, such teacher-student architecture has been employed to encode the target and context, respectively. A key difference is that the model is conditioned on actions \cite{assran2023self, garrido2024learning}, allowing free exploration in the environment. V-JEPAs \cite{bardes2024revisiting, assran2025v} encode spatiotemporal information in videos and use robot actions as the condition. CheXWorld \cite{yue2025chexworld} uses the relative location of local crops as actions to characterize the dependencies between different crops on 2D CXRs. Different from these works, our method devises an action to rotate the X-ray source to obtain projections under various transformations, allowing our world model to learn an internalized model of 3D anatomical structure that accurately predicts new projection views.

\subsection{Radiograph Visual Representation Learning}
Visual representation learning is central to developing CXR foundation models, among which vision foundation models \cite{xiao2023delving, yangchest, perez2025exploring, yue2025chexworld} learn generalizable representations via self-supervised learning objectives and vision-language foundation models \cite{huang2024enhancing, lai2024carzero, chen2024chexagent, ko2025bringing} learn semantically discriminative visual representations via contrastive alignment \cite{radford2021learning, zhai2023sigmoid}. However, the above foundation models are still limited in learning representations from 2D frontal and lateral CXRs. Our method extends the scope by incorporating 3D CT volumes into pretraining and embedding their spatial information into our model.

\section{X-ray World Intelligence Network (X-WIN)}
\label{sec:method}
This paper presents X-WIN, a novel CXR world model, for visual representation learning. 
Our method encodes spatial information from CT volumes by learning to predict its 2D projections in latent representation space. The core idea is to train a model that understands how an X-ray image would change under a known transformation in 3D space through predictive sensing \cite{yang2025cambrian}, in which models learn to predict their sensory input. Specifically, we encode a given X-ray view into a latent representation and then apply a predefined 3D action, such as a rotation, to task the model with predicting the resulting projection. If the model can accurately estimate the new X-ray projection under various actions, it is a strong indication that the latent space has internalized meaningful 3D anatomical structure. We hypothesize that the ability to reason over spatial changes of 2D projections in 3D space is critical for improving the performance of downstream CXR interpretation tasks.

We describe the overall model architecture and the action design for new X-ray projection generation in Section \ref{sec:method_overview}. We then incorporate a softened contrastive alignment objective based on the affinity matrix to train the latent projection prediction task (Section \ref{sec:method_affinity}) and a structure-preserving domain adaptation loss to encourage a cohesive embedding space between simulated and real domains (Section \ref{sec:method_domain}).

\subsection{Overall Framework}
\label{sec:method_overview}
\noindent{\textbf{World Modeling Architecture.}}
Unlike conventional deep learning architectures for vision tasks \cite{caron2021emerging, he2022masked}, world model architectures are conditioned on actions that prompt the model to predict a new state of the environment. For instance, the action in a navigation world model controls the model to predict a new scene based on the provided movement and rotation of an agent \cite{bar2025navigation}. This kind of action-conditioned prediction is central to world model training, where the model is trained to encode effective representations that can predict new states of the environment. A representative architecture is the Joint-Embedding Predictive Architecture (JEPA) \cite{lecun2022path}. 

In this paper, we introduce predictive sensing to build a world model for medical image analysis, more specifically, for CXR modeling to learn a 3D latent representation from 2D projections. The overall architecture of our proposed X-WIN is shown in Fig. \ref{fig:framework}. It consists of an encoder $f_\theta$ that receives context inputs and an encoder $f_{\theta^\prime}$ that is updated by the exponential moving average and receives target inputs. It utilizes a view predictor $g_v$ conditioned on actions to predict the representations of novel projections and a mask predictor $g_m$ to reconstruct mask tokens.

\noindent{\textbf{Action Design.}}
We formulate the action in our proposed X-WIN framework as rotation of a radiation source that emits cone-beam X-rays penetrating through a patient to generate projectional images. 
Our action formulation is inspired by the idea of tomographic scanning that rotates an X-ray source to collect projections at different angular positions to capture 3D spatial information \cite{suetens2017fundamentals}.

Specifically, we input a routine X-ray projection $u_\text{context}$ in either the frontal or lateral view into the encoder $f_{\theta}$. We then randomly sample a batch of actions $\{a_{i}\}_{i=1}^{N}$ to obtain projections of the CT volume at different angular positions $\{u_i\}_{i=1}^N$ and feed them as target projections to the encoder $f_{\theta^\prime}$. The action $a_i = k\cdot \Delta\phi$ defines the relative yaw rotation angle with respect to the position to obtain $u_\text{context}$, where $\Delta\phi$ is a fixed rotation step size and $k$ defines the number of steps. A smaller $\Delta \phi$ increases sampling density. We set $\Delta \phi$ based on the ablation study in Section \ref{sec:ablation_study}. The action $a_i$ is also fed as a condition to the view predictor $g_v$ to predict projection representations at the $a_i$ angle. This design enables the model to learn how the context X-ray projection changes given various transformations. Such capability is an indication that the latent representations encode meaningful 3D structural information. 

\subsection{Affinity-guided Contrastive Alignment}
\label{sec:method_affinity}
As shown in the upper part of Fig. \ref{fig:framework}, we utilize the view predictor $g_v$ to predict a new projection given a context projection and a predefined action. Such prediction should align with the projection after the action is applied.
Specifically, given a context projection $u_\text{context}$, the encoder $f_{\theta}$ encodes latent representations that are transformed by the view predictor $g_{v}$ as: 
\begin{equation}
\mathbf{z}_i^{\text{patch}} = g_v(\text{Linear}(a_i) \oplus (f_\theta(u_\text{context}) + \text{PE})),
\end{equation}
where $\oplus$ is a concatenation operation along the sequence dimension, $\text{Linear}(\cdot)$ is a linear layer projecting a rotation angle $a_i \in \mathbb{R}$ to the hidden dimension, $\text{PE}$ is the sinusoidal positional encoding, and $\mathbf{z}_{i}^\text{patch}$ is the predicted projection representations corresponding to $a_i$.

To learn a discriminative representation space that characterizes the features of different projection views, we incorporate a contrastive learning objective to calibrate the predicted and target representations. We leverage the InfoNCE loss \cite{oord2018representation} that aligns corresponding representations and repels unpaired representations. The target projection $u_i$ obtained via the action $a_i$ is encoded to produce target representations $\mathbf{t}_i^\text{patch} = f_{\theta^\prime}(\mathbf{u}_i)$. We apply global average pooling on $\mathbf{z}_{i}^\text{patch}$ and $\mathbf{t}_{i}^\text{patch}$ and construct representation pairs $\{(\mathbf{z}_i, \mathbf{t}_i)\}_{i=1}^{N}$. The alignment loss is formulated as:
\begin{align}
\mathcal{L}_\text{InfoNCE}= -\frac{1}{N}\sum_{i=1}^{N}\sum_{j=1}^{N}\mathbf{Y}_{ij}\log p_{ij}, \label{eq:infonce} \\
p_{ij}=\frac{\exp(\text{sim}(\mathbf{z}_i,\mathbf{t}_j)/\tau)}{\sum_{l=1}^{N}\exp(\text{sim}(\mathbf{z}_i,\mathbf{t}_l)/\tau)},
\end{align}
where $p_{ij}$ is a probability that estimates the similarity between $\mathbf{z}_i$ and $\mathbf{t}_j$, $\text{sim}(\mathbf{z}_i, \mathbf{t}_j)$ computes the dot product to measure cross-entropy similarity, $\tau$ is a learnable temperature parameter, and $\mathbf{Y}_{ij}$ is a one-hot label matrix that is equal to an identity matrix. 

The above loss $\mathcal{L}_\text{InfoNCE}$ maximizes similarity between predicted and target representations of the same projection and enforces a strict repulsive relation for unpaired representations. However, such hard alignment disregards the similarities between negative pairs. In our framework, such issue is evident as projections are sampled from the same volume and therefore contain rich anatomical correspondences. To address this issue, we incorporate an affinity-based regularizer to relax the hard alignment. The affinity matrix $\mathbf{A} \in \mathbb{R}^{N\times N}$ estimates the mutual similarities between the representations of the sampled projections $\{u_i\}_{i=1}^{N}$. The elements in $\mathbf{A}$ are calculated as:
\begin{equation}
\mathbf{A}_{ij} = \frac{\exp(\text{sim}(\mathbf{t}_i,\mathbf{t}_j)/\tilde{\tau})}{\sum_{l=1}^{N} \exp(\text{sim}(\mathbf{t}_i, \mathbf{t}_l)/\tilde{\tau})}.
\end{equation}
where $\mathbf{t}_i$ is the representations of $u_i$ generated by $f_{\theta^\prime}$ and $\tilde{\tau}$ is a learnable temperature. The affinity-based regularizer is formulated by replacing $\mathbf{Y}_{ij}$ with $\mathbf{A}_{ij}$ in Equation (\ref{eq:infonce}):
\begin{equation}
\mathcal{L}_\text{affinity} = -\frac{1}{N} \sum_{i=1}^{N} \sum_{j=1}^{N} \mathbf{A}_{ij} \log p_{ij}.
\end{equation}
The final alignment loss combines $\mathcal{L}_\text{InfoNCE}$ with the regularizer $\mathcal{L}_\text{affinity}$ weighted by $\lambda_\text{affinity}$ as 
\begin{equation}
\label{eq:align}
\mathcal{L}_\text{align} = \mathcal{L}_\text{InfoNCE} + \lambda_\text{affinity} \mathcal{L}_\text{affinity}.
\end{equation}

\subsection{Structure-preserving Domain Adaptation}
\label{sec:method_domain}
Since projection images are in the simulated domain, to enhance model adaptability on downstream tasks with real CXRs, we incorporate images in the real domain for pretraining as shown in the bottom part of Fig. \ref{fig:framework}. Specifically, we perform masked image modeling (MIM) on both real and simulated CXRs for two purposes: (i) encoding contextual and local features as achieved in \cite{yue2025chexworld} and (ii) characterizing representations in the real and simulated domains to train a domain classifier for domain adaptation. During training, we sample a set of images $\{v_i\}_{i=1}^{N}$ from the real domain, in combination with the above target projections $\{u_i\}_{i=1}^{N}$ in the simulated domain. We leverage a multi-block masking strategy \cite{assran2023self} to randomly mask out patch tokens, resulting in masked images $\tilde{v}_i$ and $\tilde{u}_i$. The output of the mask predictor $g_m$ is computed by:
\begin{equation}
\label{eq:mask_pred}
\mathbf{z}_{i}^{\prime \text{mask}} = g_m(f_{\theta}(\tilde{v}_i)\oplus m_{v_i} + \text{PE}),
\end{equation}
where $\mathbf{z}_{i}^{\prime \text{mask}}$ is the reconstructed mask tokens of the real input $\tilde{v}_i$. $m_{v_i}$ denotes mask tokens. The reconstructed mask tokens of the simulated input $\mathbf{z}_{i}^{* \text{mask}}$ is computed by replacing Equation (\ref{eq:mask_pred}) with $\tilde{u}_i$ and $m_{u_i}$. The MIM loss on real and simulated inputs is computed by:
\begin{equation}
\mathcal{L}_\text{MIM} = \frac{1}{N}\sum_{i=1}^{N}\|\mathbf{z}_{i}^{\prime \text{mask}} - \mathbf{t}_{i}^{\prime \text{mask}} \|_{2}^{2} + \frac{1}{N}\sum_{i=1}^{N}\|\mathbf{z}_{i}^{* \text{mask}} - \mathbf{t}_{i}^{* \text{mask}} \|_{2}^{2},
\end{equation}
where $\mathbf{t}_{i}^{\prime \text{mask}}$ and $\mathbf{t}_{i}^{* \text{mask}}$ are the ground-truth mask tokens of the real and simulated inputs, respectively. 

As the representations under the supervision of $\mathcal{L}_\text{MIM}$ capture specific local and contextual features, we utilize them to train a domain classifier $f_c$ for domain adaptation. We input $\mathbf{z}_{i}^{\prime \text{patch}}$ and $\mathbf{z}_{i}^{* \text{patch}}$ that contain both unmasked tokens and reconstructed mask tokens to $f_c$. The domain classifier $f_c$ contains two transformer blocks with a global average pooling layer and a linear layer on top to output logits. To obtain the capability of differentiating domain representations, $f_c$ is trained with a cross-entropy loss:
\begin{equation}
\mathcal{L}_\text{cls}=-\frac{1}{N}\sum_{i=1}^{N}(\log f_c(\mathbf{z}_{i}^{\prime \text{patch}})+\log(1-f_c(\mathbf{z}_{i}^{* \text{patch}}))).
\end{equation}
Such loss enforces the real-domain representations $\mathbf{z}_{i}^{\prime \text{patch}}$ to be associated with a probability close to one, while the simulated-domain representations $\mathbf{z}_i^{* \text{patch}}$ are with a probability close to zero.

We utilize the classifier $f_c$ to encourage a cohesive representation space that reduces the gap between real and simulated domains. Specifically, the view predictor output representations $\mathbf{z}_{i}^\text{patch}$ are enforced to be statistically similar to the real-domain representations such that they can be labeled as real by $f_c$, while $\mathbf{z}_{i}^\text{patch}$ should also retain structural information consistent with the target representations $\mathbf{t}_{i}^\text{patch}$. This is achieved by minimizing the following function \cite{tzeng2017adversarial}:
\begin{equation}
\mathcal{L}_\text{domain} = \frac{1}{N} \sum_{i=1}^{N}\|\mathbf{z}_i^\text{patch}-\mathbf{t}_i^\text{patch}\|_2^2 - \frac{1}{N}\sum_{i=1}^{N}\log f_c(\mathbf{z}_i^\text{patch}),
\end{equation}
where the first term ensures the similarity of the predicted and target patch representations and the second term updates $\mathbf{z}_i^{\text{patch}}$ such that it can be labeled as real with a probability close to one by the classifier $f_c$.

\subsection{Overall Loss}
The overall loss function consists of four parts, namely $\mathcal{L}_\text{align}$, $\mathcal{L}_\text{MIM}$, $\mathcal{L}_\text{domain}$, and $\mathcal{L}_\text{cls}$, weighted by $\lambda_\text{MIM}$, $\lambda_\text{domain}$, and $\lambda_\text{cls}$, respectively. The total loss is formulated as:
\begin{equation}
\mathcal{L_\text{overall}} = \mathcal{L}_\text{align} + \lambda_\text{MIM}\mathcal{L}_\text{MIM} + \lambda_\text{domain} \mathcal{L}_\text{domain} + \lambda_\text{cls} \mathcal{L}_\text{cls}.
\label{eq:overall}
\end{equation}
We apply a stop-gradient operation such that $\mathcal{L}_\text{domain}$ does not update $f_c$ to find a shortcut solution. We tune the weighting factors based on the ablation study in Section \ref{sec:ablation_study}.

\section{Experiments and Results}
\label{sec:experiments}
\subsection{Experimental Settings}


\noindent\textbf{Datasets.}
We trained X-WIN with 371,951 frontal and lateral CXRs from MIMIC-CXR \cite{johnson2019mimic} and 32,371 chest CT scans from the NLST \cite{flores2017ma03}. We generated X-ray projections from CT volumes with cone-beam geometry using the DiffDRR algorithm \cite{gopalakrishnan2022fast}, which supports efficient generation with GPUs and CUDA. We evaluated the model performance on six commonly used CXR benchmarks: VinDr \cite{nguyen2022vindr}, CheXpert \cite{irvin2019chexpert}, NIH-CXR \cite{wang2017chestx}, RSNA \cite{rsna}, JSRT \cite{shiraishi2000development}, and COVIDx \cite{pavlova2022covidx}. 

\noindent\textbf{Models.}
We compared the proposed X-WIN against cutting-edge models for CXR interpretation, namely the general domain self-supervised models (I-JEPA \cite{assran2023self} and DINOv2 \cite{oquab2024dinov2}), CXR foundation models (RAD-DINO \cite{perez2025exploring}, CheXFound \cite{yangchest}, Ark+ \cite{ma2025fully}, and CheXWorld \cite{yue2025chexworld}), and vision-language foundation models specialized in CXR (CARZero \cite{lai2024carzero}, CXR-Align \cite{ko2025bringing}, MaCo \cite{huang2024enhancing}, and CheXAgent \cite{chen2024chexagent}). 
For CXR foundation models, we conducted linear probing or fine-tuning of the exponential moving average encoder. For vision-language foundation models, we only used their vision encoders. 

\noindent\textbf{Training Details.}
We used ViT \cite{dosovitskiyimage} for the encoders and predictors. We set the patch size to 16 and input resolution to 512$^2$ pixels. We employed a multiblock masking strategy for masked image modeling following \cite{yue2025chexworld}.
We constrained the rotation angle $k\cdot \Delta\phi$ within $[-90^\circ, 90^\circ]$ and set the number of randomly sampled projections $N$ to 8.  
AdamW optimizer was used to train X-WIN with a batch size of 12, a cosine learning rate schedule from 2e-5 to 1e-6, a cosine weight decay schedule from 0.04 to 0.4, a cosine teacher momentum schedule from 0.994 to 1.0. The model is trained on 8$\times$ A100 40GB GPUs for 100 epochs with the number of iterations per epoch being 2,500.  More details are in the appendix. \textit{The code is available at: \url{https://github.com/RPIDIAL/X-WIN}.}

\begin{table*}[t]
    \centering
    \footnotesize
    \setlength{\tabcolsep}{8pt}
    \caption{Comparison of X-WIN with state-of-the-art models via linear probing. Results are reported in AUROC.}
    \vspace{-6pt}
    \label{tab:lin_probe}
    \begin{tabular}{l l l | c c c c c c}
        \toprule
        \textbf{Models} & \textbf{Pretrain. data} & \textbf{Backbone} & \textbf{VinDr} & \textbf{CheXpert} & \textbf{NIH-CXR} & \textbf{RSNA} & \textbf{JSRT} & \textbf{Avg.} \\ \midrule
        I-JEPA \cite{assran2023self} & ImageNet-1K & ViT-Large & 0.770 & 0.764 & 0.692 & 0.754 & 0.699 & 0.736 \\
        DINOv2 \cite{oquab2024dinov2} & LVD-142M & ViT-Base & 0.795 & 0.776 & 0.711 & 0.798 & 0.559 & 0.728 \\ \midrule
        RAD-DINO \cite{perez2025exploring} & LVD-142M + 838K CXRs & ViT-Base & 0.795 & 0.846 & 0.821 & 0.869 & 0.623 & 0.791 \\
        CheXFound \cite{yangchest} & 987K CXRs & ViT-Large & 0.869 & \underline{0.876} & 0.829 & 0.872 & \underline{0.846} & 0.858 \\
        Ark+ \cite{ma2025fully} & 704K CXRs & Swin-Base & \underline{0.906} & \underline{0.876} & \underline{0.831} & \underline{0.893} & 0.807 & \underline{0.863} \\ 
        CheXWorld \cite{yue2025chexworld} & 448K CXRs & ViT-Base & 0.903 & 0.871 & 0.833 & 0.824 & 0.791 & 0.844 \\ \midrule
        CARZero \cite{lai2024carzero} & 377K CXRs w/ reports & ViT-Base & 0.776 & 0.852 & 0.813 & 0.833 & 0.723 & 0.800 \\
        CXR-Align \cite{ko2025bringing} & 325K CXRs w/ reports & ViT-Base & 0.793 & 0.793 & 0.810 & 0.854 & 0.734 & 0.797 \\
        MaCo \cite{huang2024enhancing} & 377K CXRs w/ reports & ViT-Base & 0.817 & 0.864 & 0.827 & 0.822 & 0.745 & 0.815 \\
        CheXAgent \cite{chen2024chexagent} & 1.07M CXRs w/ text & ViT-Large & 0.823 & 0.813 & 0.756 & 0.789 & 0.634 & 0.763 \\ \midrule
        \multirow{2}{*}{X-WIN (Ours)} & \multirow{2}{*}{372K CXRs + 32K CT volumes} & ViT-Base & 0.905 & 0.902 & 0.841 & 0.917 & 0.848 & 0.883 \\
        & & ViT-Large & \textbf{0.925} & \textbf{0.908} & \textbf{0.843} & \textbf{0.929} & \textbf{0.857} & \textbf{0.892} \\
        \bottomrule
    \end{tabular}
    \vspace{-7pt}
\end{table*}

\begin{table}[t]
    \centering
    \footnotesize
    \setlength{\tabcolsep}{8pt}
    \caption{Comparison of X-WIN with vision foundation models on few-shot fine-tuning for COVID-19 pneumonia detection. Five-run averages in AUROC are reported.}
    \vspace{-8pt}
    \scalebox{1.0}{
    \label{tab:finetuning}
    \begin{tabular}{l | c c c c}
        \toprule
        \textbf{Models} & \textbf{4 shots} & \textbf{8 shots} & \textbf{16 shots} & \textbf{All} \\ \midrule
        CheXFound & 0.823 & 0.883 & 0.897 & 0.977 \\
        RAD-DINO & 0.793 & 0.852 & 0.884 & 0.971\\
        Ark+ & \textbf{0.872} & \underline{0.908} & \underline{0.921} & \underline{0.985} \\
        CheXWorld & 0.843 & 0.893 & 0.902 & 0.981 \\ \midrule
        X-WIN (Ours) & \underline{0.868} & \textbf{0.924} & \textbf{0.939} & \textbf{0.993} \\ 
        \bottomrule
    \end{tabular}
    }
    \vspace{-2pt}
\end{table}

\subsection{Main Results}
\noindent{\textbf{Model Comparison via Linear Probing.}}
We utilized linear probing to evaluate representation quality. We froze the vision encoder and trained a learnable linear layer at the end for classification. Table~\ref{tab:lin_probe} shows the comparison of X-WIN against state-of-the-art CXR representation learning methods. Those methods are divided into three groups. 

The first group, including I-JEPA \cite{assran2023self} and DINOv2 \cite{oquab2024dinov2}, pretrains the vision encoders with general domain images. These models achieved an average AUROC of around 0.730, which is a reasonable result given that they were trained with general domain images but not CXRs. 

The second group contains CXR foundation models (RAD-DINO \cite{perez2025exploring}, CheXFound \cite{yangchest}, Ark+ \cite{ma2025fully}, and CheXWorld \cite{yue2025chexworld}). Compared to group 1, these models enhance visual representation learning by incorporating large-scale CXRs. However, they are limited to modeling structural information in frontal and lateral CXRs, whereas our proposed model additionally distills volumetric information from CT. The capability to model new 2D CXRs under transformations in 3D space led to our enhanced performance. Ark+ is the second-best performing model. It relied on diverse disease labels in pretraining to achieve discriminative representations, whereas our method alternatively employed a contrastive loss to build attractive and repulsive relations between representations and also achieved strong discriminability. 

The third group contains CXR vision-language foundation models (CARZero \cite{lai2024carzero}, CXR-Align \cite{ko2025bringing}, MaCo \cite{huang2024enhancing}, and CheXAgent \cite{chen2024chexagent}). These models generally underperformed CXR foundation models for the following reasons. CLIP-based models \cite{lai2024carzero, ko2025bringing, huang2024enhancing} still struggle with aligning intricate disease information in radiology reports with local features on CXRs. The generative model \cite{chen2024chexagent} fails to selectively attend to critical information in reports. 
It is worth noting that X-WIN outperformed the recent CXR world model CheXWorld \cite{yue2025chexworld}. CheXWorld learns local structures, global geometry, and appearance variation of 2D radiographs, whereas X-WIN focuses on internalizing 3D anatomical structures into our world model. The leap from 2D to 3D underlies our performance gains. 

\begin{figure}[t]
    \centering
    \includegraphics[width=\linewidth]{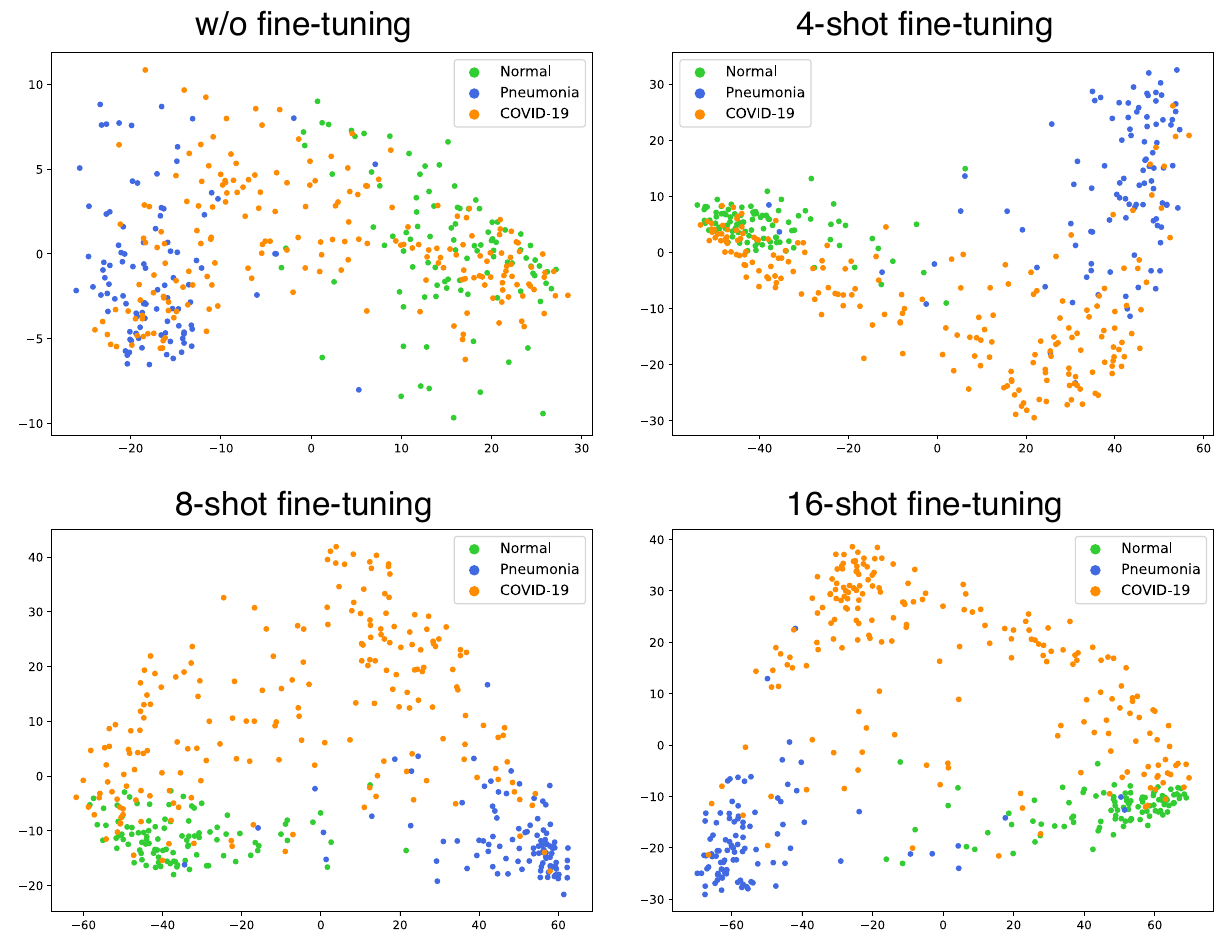}
    \vspace{-16pt}
    \caption{t-SNE visualizations of encoder representations with corresponding semantic labels. The representations of X-WIN demonstrate a modest separation between semantic classes without finetuning. With 4-shot fine-tuning, clear clusters for three classes can be observed.}
    \vspace{-8pt}
    \label{fig:fewshot_ft}
\end{figure}

\noindent{\textbf{Adaptability Evaluation with Fine-tuning.}} Besides linear probing, we evaluated the model adaptability with few-shot fine-tuning. Such adaptability is crucial in real-world applications, enabling models to adapt to new diseases and datasets. We conducted experiments on the COVIDx dataset \cite{pavlova2022covidx} that contains CXRs from normal, pneumonia, and COVID-19 patients. We sampled 4, 8, and 16 samples per class for few-shot fine-tuning. We sampled the few-shot labels with five different random seeds and reported the average AUROC over five runs. 
As shown in Table \ref{tab:finetuning}, X-WIN achieved an AUROC of 0.993, outperforming all the comparison models. As shown in the t-SNE visualizations in Fig. \ref{fig:fewshot_ft}, the fine-tuned X-WIN generated clearly separated representations for classes, demonstrating its strong discriminability and adaptability.
In the 4-shot setting, X-WIN outperformed all models except Ark+ which additionally utilized disease labels in pretraining and also learned discriminative representations. We report additional fine-tuning results on VinDr and RSNA in the appendix.

\subsection{Ablation Study}
\label{sec:ablation_study}
\begin{figure}[t]
    \centering
    \includegraphics[width=\linewidth]{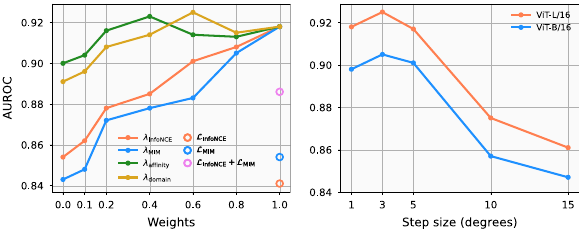}
    \vspace{-18pt}
    \caption{Hyperparameter tuning. Left: Effects of loss functions (circles) and sensitivity analysis on loss weights (lines). Right: Analysis of the impact of rotation step sizes on model performance. We reported performance on VinDr via linear probing.}
    \vspace{-3pt}
    \label{fig:hyperparams}
\end{figure}

\noindent{\textbf{Loss Functions.}}
We evaluated the effects of individual loss functions and analyzed their weights (Fig. \ref{fig:hyperparams} left). Applying $\mathcal{L}_\text{InfoNCE}$ or $\mathcal{L}_\text{MIM}$ alone achieved strong baseline performance. Combining the two losses led to substantial improvements owing to their synergistic effects, where $\mathcal{L}_\text{MIM}$ and $\mathcal{L}_\text{InfoNCE}$ were employed to encode local image features and distilled 3D spatial features from CT, respectively. 
We analyzed the sensitivity of our model to the weights of different losses by first setting a weight of 1.0 for losses in Eqs.~(\ref{eq:align}) and (\ref{eq:overall}) in the benchmark experiment and then gradually decreasing a loss weight while keeping other loss weights fixed at 1.0. Experimental results show that decreasing the loss weights to 0.0 consistently led to performance degradation, demonstrating their individual contributions to our method. We observe that tuning $\mathcal{\lambda}_\text{affinity}$ to 0.4 and $\lambda_\text{domain}$ to 0.6 achieved the best performance and set them as the default weights. $\mathcal{L}_\text{cls}$ does not update the encoder and we set its weight to 1.0. 

\noindent{\textbf{Action Design.}} We conducted experiments to analyze the effects of different action designs (Table \ref{tab:action_design}). Results show that employing step-wise rotation (yaw) underperformed direct rotation (yaw). The step-wise rotation divides the direct rotation of $k\cdot \Delta \phi$ degrees into $k$ individual steps. Such design puts emphasis on modeling transition dynamics between adjacent projections, which diminished the degree of freedom and led to degraded performance. Additionally, we conducted an experiment on direct rotation with three-dimensional Euler angles defining yaw, pitch, and roll. We observe that the setting with yaw rotation performed better than the setting with three-dimensional rotation, probably owing to the sufficient 3D spatial information contained in the projections with yaw rotation and their consistency with routine projections.

\noindent{\textbf{Rotation Step Size.}}
We evaluated the impact of rotation step size $\Delta \phi$ to downstream performance (Fig. \ref{fig:hyperparams} right). We set the rotation step size to \{1$^\circ$, 3$^\circ$, 5$^\circ$, 10$^\circ$, 15$^\circ$\}, resulting in \{180, 60, 36, 18, 12\} potential projections for sampling. Experimental results show that a 3$^\circ$ rotation step size achieved the highest performance. Increasing the rotation step size to \{5$^\circ$, 10$^\circ$, 15$^\circ$\} substantially degraded the performance due to increased sparsity. Our experiment with a 1$^\circ$ step size underperformed the 3$^\circ$ setting likely due to the reduced diversity of targets as the step size decreases.

\begin{table}[t]
    \centering
    \footnotesize
    \setlength{\tabcolsep}{10pt}
    \caption{Analysis of the impact of different action designs. We report linear probing results in AUROC.}
    \vspace{-8pt}
    \label{tab:action_design}
    \begin{tabular}{l | c c}
    \toprule
    \textbf{Action design} & \textbf{VinDr} & \textbf{CheXpert}  \\ \midrule
    Step-wise rotation (yaw)  & 0.857 & 0.843 \\
    Direct rotation (yaw)  & \textbf{0.925} & \textbf{0.908} \\
    Direct rotation (yaw, pitch, and roll) & 0.871 & 0.865 \\
    \bottomrule
    \end{tabular}
    \vspace{-3pt}
\end{table}



\begin{figure}
    \centering
    \includegraphics[width=\linewidth]{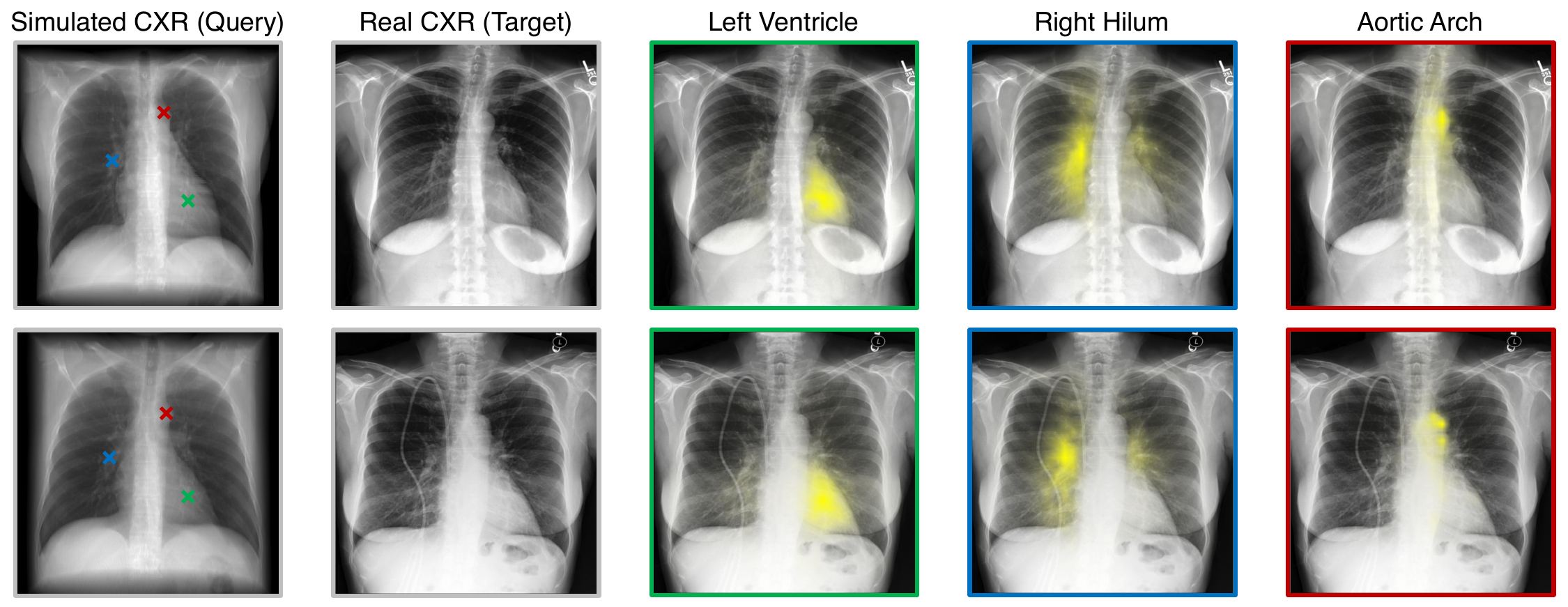}
    \vspace{-16pt}
    \caption{Analysis of representation correspondences. We compute cosine similarities between the anatomical landmarks in the simulated CXRs and the representations of the real CXRs. Regions highlighted in yellow indicate patch representations with high similarities.}
    \vspace{-8pt}
    \label{fig:patch_correspondences}
\end{figure}

\noindent{\textbf{Effect of Domain Adaptation.}}
We quantified the similarity between simulated-domain and real-domain representations with/without applying $\mathcal{L}_\text{domain}$. We computed the cluster centers of image representations in the held-out test set on NLST and the official test set on VinDr, and then measured their similarity. Table \ref{tab:domain_adaptation} shows that applying $\mathcal{L}_\text{domain}$ improved cosine similarity and reduced L2 distance owing to the domain classifier's effect in enforcing statistical-level alignment. Fig. \ref{fig:patch_correspondences} shows that our model built patch-level correspondences between domains owing to local features learned via MIM in both domains and the domain adaptation loss.

\begin{table}[t]
    \centering
    \footnotesize
    \caption{Quantitative analysis of the effect of the domain adaptation loss. We report the similarity values between the representations of simulated and real CXRs.}
    \vspace{-8pt}
    \begin{tabular}{c|c c}
    \toprule
       \textbf{Pretrain. method}  & \textbf{Cosine similarity} & \textbf{L2 distance} \\ \midrule
        w/o $\mathcal{L}_\text{domain}$ & 0.845 & 3.973 \\
        w/ $\mathcal{L}_\text{domain}$ & 0.967 & 1.825 \\
    \bottomrule
    \end{tabular}
    \label{tab:domain_adaptation}
\end{table}

\subsection{Reconstruction of 3D CT Volumes}
\label{sec:reconstruction}
\begin{figure}[t]
    \centering
    \includegraphics[width=0.92\linewidth]{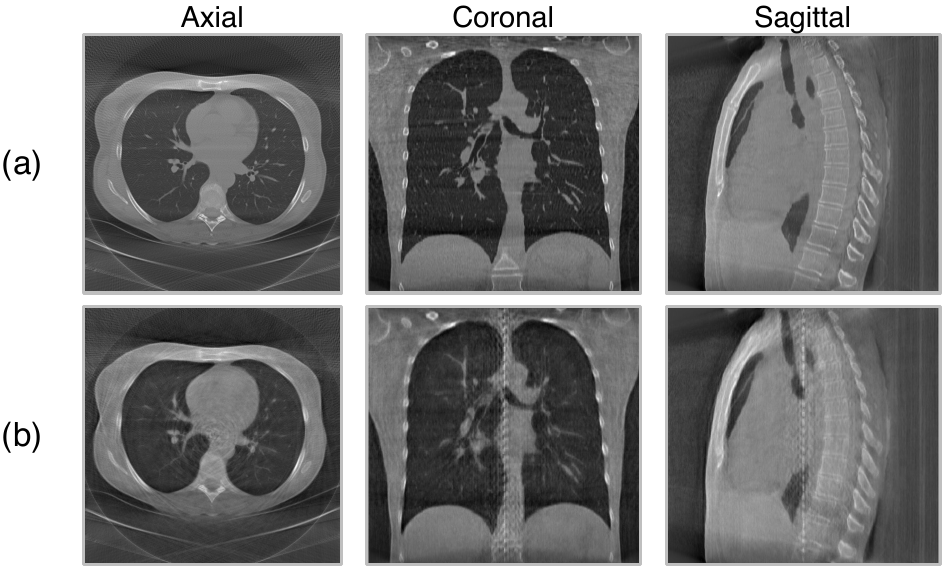}
    \caption{Reconstructed CT volumes in three views. (a) and (b) show reconstruction results with ground-truth projections and rendered projections, respectively. Our method preserves the overall shape and a reasonable degree of low-level structural information. We observe a certain level of blurriness probably due to the abstraction of visual representation learning.}
    \label{fig:recon_slices}
    \vspace{-8pt}
\end{figure}

In line with our objective of volumetric knowledge distillation from CT, we conducted experiments to verify whether the learned representations capture 3D information.
We trained a VQ-GAN decoder \cite{esser2021taming} on top of the frozen encoder $f_{\theta}$ and view predictor $g_{v}$. We trained the decoder with ground-truth projections. At inference time, we utilized the FDK algorithm in the ASTRA toolbox \cite{van2016fast} to reconstruct a 3D CT volume from rendered projections. More details are described in the appendix.
To optimize projection rendering performance, we varied the codebook size and codebook vector dimension following \cite{sun2024autoregressive}. The trained decoder attained a high codebook usage of 81.6\%, which is calculated as the percentage of used codes at inference time. The rendered 2D projections achieved a peak signal-to-noise ratio (PSNR) of 30.23 dB and structural similarity index measure (SSIM) of 0.888. The reconstructed 3D volumes achieved a PSNR of 27.87 dB and SSIM of 0.789. Fig. \ref{fig:recon_slices} demonstrates that the reconstructed volume preserves the overall structure and a reasonable level of local details while a certain level of loss in local details occurs due to the abstraction of visual representations.


\begin{table}[t]
    \centering
    \setlength{\tabcolsep}{7pt}
    \footnotesize
    \caption{Results on 3D CT reconstruction. The VQ-GAN decoder is trained with different codebook sizes and dimensions. We evaluate performance on rendered projections and reconstructed CT.}
    \vspace{-8pt}
    \label{tab:recon}
    \begin{tabular}{c c c | c c | c c}
        \toprule
        \multicolumn{3}{c|}{\textbf{Codebook}} & \multicolumn{2}{c|}{\textbf{Projection}} & \multicolumn{2}{c}{\textbf{CT volume}} \\
        \textbf{Size} & \textbf{Dim.} & \textbf{Usage} & \textbf{PSNR} & \textbf{SSIM} & \textbf{PSNR} & \textbf{SSIM} \\ \midrule
        1024 & 128 & 73.9\% & 24.77 & 0.743 & 20.89 & 0.686 \\
        2048 & 128 & 44.5\% & 25.79 & 0.759 & 21.57 & 0.702 \\
        1024 & 256 & 81.6\% & \textbf{30.23} & \textbf{0.888} & \textbf{27.87} & \textbf{0.789} \\
        2048 & 256 & 78.8\% & 27.50 & 0.783 & 24.32 & 0.723 \\
        \bottomrule
    \end{tabular}
\end{table}




\section{Discussion and Conclusion}
\label{sec:conclusion}

This work presents the X-WIN framework, a novel world model that effectively learns 3D spatial information from CT volumes for application to CXRs. By projecting CT volumes into simulated CXR views, X-WIN enables volumetric knowledge distillation and leverages a softened contrastive loss to achieve robust cross-view alignment. The model adapts to real CXRs through a statistical domain adaptation loss, building a cohesive embedding space between simulated and real domains. Our experiments demonstrate that utilizing simulated CXRs with contrastive alignment establishes a strong baseline, while incorporating real data and domain adaptation further enhances performance.

The proposed X-WIN shows promising results, outperforming recent methods across diverse downstream tasks and learning representations that are interpretable via 3D reconstruction. Our future work points to two directions. On the technical side, we aim to further address the sim-to-real domain adaptation challenge. For instance, exploring invariant feature learning or domain mapping strategies \cite{tobin2017domain} may further improve data efficiency and model generalizability. On the experimental side, we plan to compare our model performance to models trained directly on CT, which will help demonstrate the full potential of CXRs for disease detection and diagnosis. Overall, this work provides an important step toward integrating 3D anatomical knowledge into 2D imaging models, paving the way for more accurate and explainable AI systems in radiographic analysis.

\section*{Acknowledgments.}
This work was supported by the National Science Foundation through the Faculty Early Career Development Program (CAREER) under Award 2046708 and the National Institutes of Health's National Institute on Aging under Grant T32AG078123.

{
    \small
    \bibliographystyle{ieeenat_fullname}
    \bibliography{main}
}


\end{document}